\documentclass[letterpaper, 10 pt, conference]{ieeeconf}  

\IEEEoverridecommandlockouts                              

\usepackage{graphicx} 
\usepackage{mathptmx} 
\usepackage{times} 
\usepackage{amsmath} 
\usepackage{amssymb}  
\usepackage{textcomp}
\usepackage{booktabs}
\usepackage{multirow}
\usepackage{balance}
\usepackage{soul,color}
\usepackage{comment}
\usepackage{array}
\usepackage[ruled,vlined]{algorithm2e}
\usepackage{cite}
\usepackage{etoolbox}
\usepackage{tikz}
\usepackage{cuted}
\usepackage{capt-of}
\makeatletter
\let\NAT@parse\undefined
\makeatother

\newcommand{\pizerofive}{\ensuremath{\pi_{0.5}}}

\makeatletter
\patchcmd{\@makecaption}
  {\begin{center}{\footnotesize #1}\\{\footnotesize\scshape #2}\end{center}}
  {%
    \par\noindent
    \begingroup
    \fontsize{8}{9}\selectfont
    \setlength{\baselineskip}{9pt}%
    \textbf{#1:}~#2\par
    \endgroup
  }
  {}
  {\PackageWarning{tablecaption}{Table caption patch failed}}
\makeatother

\usepackage[
    colorlinks=true,
    linkcolor=blue,
    citecolor=blue,
    urlcolor=blue
]{hyperref}

\makeatletter
\@ifundefined{bstctlcite}{%
  \def\bstctlcite{\@ifnextchar[{\@bstctlcite}{\@bstctlcite[@auxout]}}%
  \def\@bstctlcite[#1]#2{\@bsphack
    \@for\@citeb:=#2\do{%
      \edef\@citeb{\expandafter\@firstofone\@citeb}%
      \if@filesw\immediate\write\csname #1\endcsname{\string\citation{\@citeb}}\fi}%
    \@esphack}%
}{}
\makeatother

\title{\LARGE 
RESCUE: Is This Alarm Worth Acting On? \\ Counterfactual Evaluation of VLA Runtime Monitors
}

\author{David Ahmedt-Aristizabal,
Abdelwahed Khamis,
Mohammad Ali Armin, 
Lars Petersson \\
CSIRO, Australia\\ 
{\tt\small \{David.Ahmedtaristizabal,~Lars.Petersson\}@csiro.au}
}

\begin{document}

\bstctlcite{IEEEexample:BSTcontrol}








\twocolumn[{%
\renewcommand\twocolumn[1][]{#1}%
\maketitle
\thispagestyle{empty}
\pagestyle{empty}
\begin{minipage}{\textwidth}
    \centering
    \vspace{-0.8cm}
    \includegraphics[width=0.65\textwidth]
    {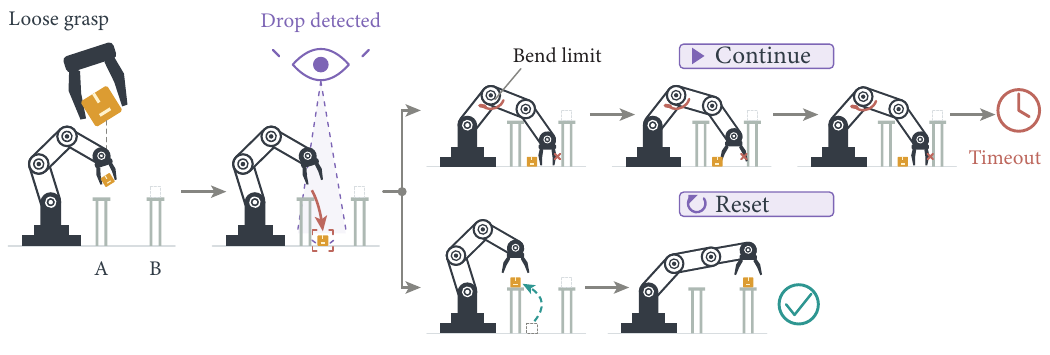}

    \vspace{-0.4cm}

    \captionof{figure}{
    \textbf{Failure prediction does not necessarily imply intervention value.}
    RESCUE evaluates an existing runtime-monitor alarm by comparing counterfactual interventions from the same robot-policy state.
    Here, a loose grasp drops the block into a difficult-to-reach location: continuation leads to repeated unsuccessful attempts and timeout, whereas resetting completes the task earlier despite its overhead.
    Intervention value therefore depends on the available response, alarm timing, false alarms, failure prevalence, and intervention cost.
    }
    \label{fig:rescue_overview}

\end{minipage}

\vspace{3mm}
}]

\begin{abstract}
Runtime failure monitors for robot policies are typically evaluated by how accurately they distinguish failed from successful executions.
Yet an accurate alarm may arrive too late for recovery, while an unnecessary intervention may disrupt an execution that would otherwise succeed.
We introduce RESCUE, a counterfactual framework for evaluating whether a monitor's alarms enable useful interventions.
By restoring simulator and policy states at execution checkpoints, RESCUE compares alternative actions, identifies the latest checkpoint satisfying a specified recovery criterion, and evaluates whether acting on alarms reduces expected task-completion cost.
We evaluate RESCUE with OpenVLA-OFT and \pizerofive{} over LIBERO-Spatial and LIBERO-Plus using 128,340 counterfactual branches.
Recoverability depends on the state--intervention pair rather than on robot state alone: at $\tau=0.5$, 14 of 66 OpenVLA-OFT/LIBERO-Plus failures contain a recoverable checkpoint under continuation, compared with 34 of 66 under continuation or scripted recovery, while recovery reduces recoverability in other conditions.
On the primary strong-policy conditions, favourable detection scores do not establish lower intervention cost.
Under the evaluated stop-and-reset model, even an idealised detector can afford only a 2.0--3.3\% false-alarm rate before unnecessary resets outweigh the savings from stopping failures early.
These results show that monitor value depends jointly on the intervention available and the workload in which it is deployed.

\end{abstract}

\vspace{-0.2cm}
\section{INTRODUCTION}
%
%


Vision-language-action (VLA) models are increasingly capable of executing diverse robotic manipulation tasks from visual observations and language instructions~\cite{kim2025openvlaoft,physicalintelligence2025pi05}. 
Despite this progress, long-horizon execution remains vulnerable to compounding errors, environmental perturbations, and out-of-distribution conditions. 
Runtime monitors therefore aim to detect impending or ongoing failures from policy representations, observations, or predictive world models so that execution can be stopped, recovered, replanned, or escalated~\cite{gu2025safe,seligmann2026vlafail,liu2026checkvla,zhang2026foresight}. 
These monitors are commonly selected using discrimination, calibration, or timing-aware detection metrics.

However, predicting failure and deciding whether to intervene are different problems. 
A monitor may accurately distinguish failed from successful executions while still providing a poor intervention rule: an alarm may arrive after useful recovery has disappeared, or a false alarm may interrupt an execution that would otherwise succeed. 
Conversely, an imperfect predictor may still be useful if it alarms while an effective and inexpensive intervention remains available. 
Related agent studies similarly show that better failure prediction or calibration need not improve outcomes under intervention~\cite{vasudev2026failure,zhang2026calibration}. 
In robotics, this distinction is particularly important because interventions alter physical state and may incur reset, delay, or damage costs.

Timing-aware evaluation addresses part of this problem, but elapsed time alone is insufficient. VLA-FAIL~\cite{seligmann2026vlafail} incorporates detection time into monitor evaluation, PREFAIL~\cite{shangguan2026prefail} identifies an actionable intervention window for lift-and-place failures, and CheckVLA~\cite{liu2026checkvla} defines an intervention-relative irreversibility time by testing whether a fixed set of repairs can still restore success.
CheckVLA establishes that actionability depends on available repairs, but leaves open how intervention value changes across the trajectory and whether acting remains preferable once false alarms and intervention costs are included~\cite{liu2026checkvla}.
The same robot-policy state can therefore have different value under continuation, recovery, or stop-and-reset.

We introduce \textbf{RESCUE} (\textbf{R}ecoverability \textbf{E}valuation of \textbf{S}topping \textbf{C}riteria \textbf{U}nder \textbf{E}xecution), a counterfactual framework for evaluating existing runtime monitors as intervention rules. 
As illustrated in Fig.~\ref{fig:rescue_overview}, RESCUE restores simulator and policy states at action-chunk boundaries and evaluates alternative interventions, estimating action-conditioned task success under a declared intervention repertoire $\mathcal{A}$, execution model $\xi$, and remaining task budget $B$.
Evaluating these outcomes across the trajectory yields a \emph{recoverability horizon}: the latest checkpoint at which the declared repertoire satisfies a specified recovery criterion. 
The horizon therefore depends explicitly on the available interventions, execution model, and task budget.

Recoverability establishes what can still succeed, but not whether intervention is worthwhile. Stopping a failed execution can save time, whereas an unnecessary stop incurs a reset and another attempt.
RESCUE therefore evaluates interventions in a common cost-to-completion space incorporating failure prevalence, alarm timing, false alarms, and intervention cost.
For stop-and-reset, this yields a closed-form break-even false-alarm rate computable from ordinary execution logs, allowing a monitor's realised operating point to be compared directly with never intervening.

We evaluate RESCUE with OpenVLA-OFT and $\pi_{0.5}$ on LIBERO-Spatial and LIBERO-Plus~\cite{liu2023libero,fei2025liberoplus}, comprising 1,562 trunk episodes and 128,340 counterfactual branches. 
At $\tau=0.5$, only 14 of 66 OpenVLA-OFT/LIBERO-Plus failures contain a recoverable checkpoint under continuation, while 34 of 66 clear under continuation or scripted recovery. 
Under stop-and-reset, even an idealised detector that identifies every failure immediately can afford only a 2.78--3.32\% false-alarm rate under the evaluated cost model. 
A steps-since-progress monitor tuned to a nominal 5\% operating point instead costs $+0.0835$ reset-equivalents per completed task relative to never intervening. 
Across matched evaluation sets, conventional detection and timing metrics admit the real monitors in 61 of 75 verdicts, whereas decision value admits none of 15 against never intervening. 
These results show that failure-prediction quality alone does not establish intervention value.
Our contributions are:
\begin{enumerate}
    \item We formulate runtime-monitor evaluation over \emph{state--intervention pairs} and introduce a counterfactual branching protocol that estimates intervention success along robot-policy trajectories, yielding a recoverability horizon conditional on the intervention repertoire, execution model, and remaining task budget.

    \item We complement recoverability with a decision-value formulation accounting for failure prevalence, alarm timing, false alarms, and intervention cost, and derive a closed-form break-even false-alarm criterion for stop-and-reset from standard execution logs.

    \item We evaluate the framework at scale across two VLA policies and show that recoverability changes substantially with the available intervention, while monitors favoured by conventional detection and timing metrics can remain decision-negative when evaluated against the intervention they trigger.
\end{enumerate}

\section{RELATED WORK}
\label{sec:related_work}


\textit{Runtime Failure Monitoring for Robot Policies.}
Runtime monitors detect execution failures using policy representations, action consistency, and predictive world models.
Sentinel~\cite{agia2025sentinel} pairs temporal action consistency with VLM-based task-progress reasoning.
FIPER~\cite{romer2025fiper} combines embedding-space out-of-distribution detection with action-chunk uncertainty, conformally calibrating both scores on successful rollouts.
For VLA policies, SAFE~\cite{gu2025safe} predicts failures from policy representations, VLA-FAIL~\cite{seligmann2026vlafail} introduces AUCPDT to assess detection accuracy and timing, and Foresight~\cite{zhang2026foresight} uses action-conditioned world-model latents to detect long-horizon manipulation failures.
These works primarily evaluate how reliably and how early failures are identified.
CheckVLA~\cite{liu2026checkvla} couples detection to repair by evaluating suffix rewrites from restored simulator states.
It defines an \emph{irreversibility time} beyond which none of its tested repairs restores success, and reports rescue, harm, and task success.
RESCUE instead compares alternative interventions throughout execution and evaluates whether acting on alarms lowers expected task-completion cost under the workload's failure prevalence and intervention costs.


\textit{Failure Recovery and Intervention Timing.}
Recovery methods address how a robot should respond when execution goes wrong.
AEGIS~\cite{chen2026aegis} selectively escalates to a stronger policy,
B2FF~\cite{shin2026b2ff} selects a pre-imagined familiar future as a recovery goal, and
Rewind-IL~\cite{zheng2026rewind} returns the robot toward a previously verified safe state.
PREFAIL~\cite{shangguan2026prefail} identifies the latest actionable intervention time for lift-and-place failures, assessing whether predictions arrive early enough to prevent failure.
These works focus on designing recovery behaviour or assessing whether intervention occurs within an actionable window.
RESCUE instead evaluates declared branchable interventions from every saved action-chunk boundary, allowing recovery opportunities to be measured throughout execution rather than only at a single trigger point.
The recoverability estimate is conditional on the intervention repertoire, execution model, and remaining task budget.
RESCUE therefore defines the recoverability horizon as the latest checkpoint satisfying the recovery criterion, without assuming that recoverability decreases monotonically with execution time.

\begin{figure*}[!t]
    \centering
    \includegraphics[width=0.8\linewidth]{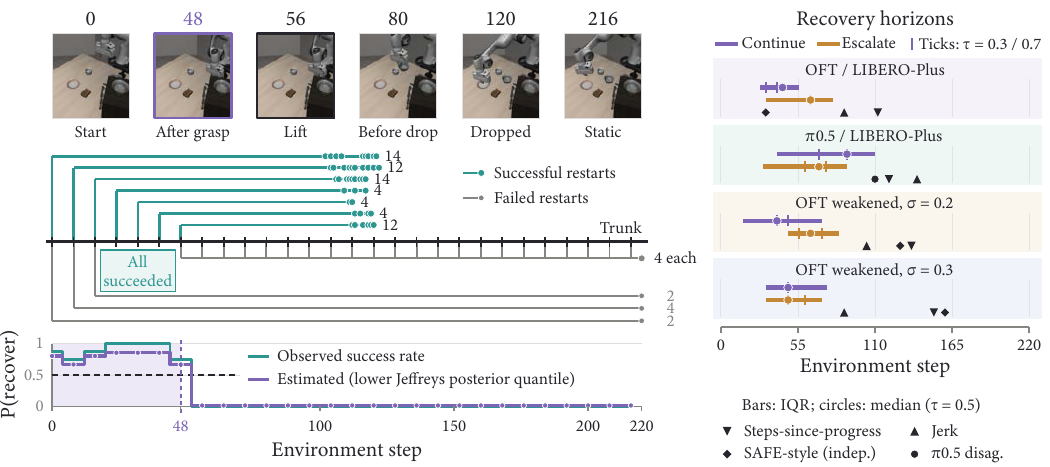}
    \vspace{-0.3cm}
    \caption{\textbf{Measuring recoverability and its dependence on intervention.}
\textbf{Left:} Counterfactual restarts from saved checkpoints of a failed OpenVLA-OFT/LIBERO-Plus execution yield observed success rates (teal) and lower-quantile recoverability estimates (purple). At $\tau=0.5$, step 48 is the last recoverable checkpoint, while step 56 no longer satisfies the recovery criterion.
\textbf{Right:} Median recoverability horizons (circles), interquartile ranges (bars), and median runtime-monitor alarm times (black symbols) across policy settings. Continue and escalate denote \textsc{Continue} and scripted \textsc{Recover}; ticks show median horizons at $\tau=0.3$ and $\tau=0.7$. On OpenVLA-OFT/LIBERO-Plus, 14/66 failures contain a qualifying checkpoint under \textsc{Continue}, 33/66 under \textsc{Recover}, and 34/66 under their union.
}
    \label{fig:rescue_horizon_example}
\vspace{-0.4cm}
\end{figure*}

\textit{From Prediction Quality to Decision Value.}
Predictive accuracy does not guarantee beneficial decisions, a distinction reflected in cost-sensitive learning~\cite{elkan2001foundations} and critiques of aggregate evaluation measures~\cite{vickers2006decision,drummond2006cost,hand2009auc}.
ThriftyDAgger~\cite{hoque2021thriftydagger} studies intervention budgets in interactive imitation learning.
For LLM agents, Vasudev et al.~\cite{vasudev2026failure} show that interventions can both rescue failed executions and disrupt successful ones.
Zhang et al.~\cite{zhang2026calibration} measure \emph{intervention advantage} by branching alternative actions from the same execution prefix, showing that better risk calibration need not reduce control regret.
RESCUE brings this decision-oriented view to robot manipulation, where interventions alter physical state and resets require another attempt.
It evaluates existing monitors by comparing counterfactual interventions from restored robot-policy states and measuring expected cost per completed task under a reset-and-retry model.
Rather than computing a model-based reachable set~\cite{bansal2017reachability}, RESCUE estimates action-conditioned recoverability empirically from restored execution states.
For stop-and-reset, this yields a break-even false-alarm rate computable from ordinary execution logs.

\section{METHOD}

RESCUE evaluates an existing runtime monitor by asking what each available intervention could have achieved from the robot-policy state at which the monitor alarms. 
An ordinary policy execution forms a \emph{trunk} trajectory, from which complete robot-policy states are saved at discrete checkpoints. 
For a checkpoint $s_t$, RESCUE restores the same state and evaluates alternative interventions through counterfactual continuations. 
Each state--intervention pair is characterised by two complementary quantities: the probability of completing the current task within the remaining budget, and the expected cost to eventual task completion. 
The former determines whether a useful intervention remains available, whereas the latter determines whether acting is preferable once reset, delay, and retry costs are considered.

These quantities are conditional on a declared intervention repertoire $\mathcal{A}$, execution model $\xi$, and remaining task budget $B$. 
Evaluating success probability across checkpoints yields an action-conditioned \emph{recoverability horizon}, while evaluating interventions in a common cost space allows monitor alarms to be assessed as decision rules rather than solely as failure predictions. 
Fig.~\ref{fig:rescue_horizon_example} illustrates how the recoverability horizon is measured on one failed execution by repeatedly restoring checkpoints and observing their counterfactual outcomes.

\subsection{State--Intervention Value and Recoverability}
\label{sec:recoverability}

Let $\pi$ denote the nominal robot policy and let $s_t$ be the complete restorable robot-policy state at checkpoint $t\in\mathcal{T}$. 
For a branchable intervention $a$ applied at $s_t$, let $\xi$ denote the execution model governing subsequent stochasticity and let $B(t)$ denote the remaining budget of the current attempt.

We characterise each state--intervention pair through two complementary quantities. 
First, its within-attempt success probability is
\vspace{-0.2cm}
\begin{equation}
Q^{\mathrm{succ}}_{\xi,B}(s_t,a)
=
\Pr_{\xi}
\!\left[
\text{success within } B(t)
\mid s_t,a,\pi
\right].
\label{eq:q_success}
\end{equation}
Second, its expected cost to eventual task completion is
\begin{equation}
Q^{\mathrm{cost}}_{\xi,B}(s_t,a)
=
\mathbb{E}_{\xi}
\!\left[
C
\mid s_t,a,\pi
\right],
\label{eq:q_cost}
\end{equation}
where $C$ includes the cost incurred by the current attempt and, if needed, subsequent reset-and-retry attempts.

Let $\mathcal{A}$ denote the repertoire of branchable interventions available at the checkpoint. 
The best success probability and minimum expected cost available from $s_t$ are
\begin{equation}
\small
V_{\mathcal{A},\xi,B}(s_t)
=
\max_{a\in\mathcal{A}}
Q^{\mathrm{succ}}_{\xi,B}(s_t,a),
\quad
U_{\mathcal{A},\xi,B}(s_t)
=
\min_{a\in\mathcal{A}}
Q^{\mathrm{cost}}_{\xi,B}(s_t,a).
\label{eq:repertoire_values}
\end{equation}

For a recovery criterion $\tau\in[0,1]$, we call $s_t$ \emph{recoverable} when
$V_{\mathcal{A},\xi,B}(s_t)\geq\tau$.
The corresponding \emph{recoverability horizon} is the latest checkpoint satisfying this criterion,
\begin{equation}
H_{\mathcal{A},\xi,B,\tau}
=
\max
\left\{
t\in\mathcal{T}:
V_{\mathcal{A},\xi,B}(s_t)\geq\tau
\right\},
\label{eq:recoverability_horizon}
\end{equation}
and is undefined when no checkpoint satisfies the criterion.

This formulation makes recoverability a property of the state--intervention pair rather than of the robot state alone. 
Changing the intervention repertoire, execution model, or remaining budget can therefore change both whether a checkpoint is recoverable and where the horizon lies. 
For a monitor alarm at checkpoint $\hat{t}$, $V_{\mathcal{A},\xi,B}(s_{\hat{t}})$ describes what remains achievable within the current attempt, while $U_{\mathcal{A},\xi,B}(s_{\hat{t}})$ provides the corresponding decision-oriented view in cost units. 
The horizon does not assume that recoverability varies monotonically with time.

\subsection{Counterfactual Branching and Estimation}
\label{sec:counterfactual_estimation}

RESCUE estimates the state--intervention quantities in
Sec.~\ref{sec:recoverability} by repeatedly restoring the same checkpoint and rolling execution forward under a specified intervention.
A trunk episode is executed normally while complete robot-policy states are saved at action-chunk boundaries.
For checkpoint $s_t$ and branchable intervention $a\in\mathcal{A}$, the state is restored and $K_{t,a}$ seeded counterfactual branches are generated under execution model $\xi$, each terminating at task success or when the remaining budget $B(t)$ is exhausted.
The same branches provide both success outcomes and realised execution lengths for estimating
$Q^{\mathrm{succ}}_{\xi,B}(s_t,a)$ and
$Q^{\mathrm{cost}}_{\xi,B}(s_t,a)$.

Let $k_{t,a}$ be the number of successful branches.
Because the initial branch count is small, we estimate success probability using a lower Jeffreys posterior quantile rather than the raw fraction,
\begin{equation}
\widehat{Q}^{\mathrm{succ}}_{\xi,B}(s_t,a)
=
Q_q
\!\left[
\mathrm{Beta}
\left(
k_{t,a}+\tfrac{1}{2},
K_{t,a}-k_{t,a}+\tfrac{1}{2}
\right)
\right],
\label{eq:q_success_estimator}
\end{equation}
where $Q_q[\cdot]$ denotes the $q$-quantile and $q=0.25$.
Each checkpoint is initially evaluated with $K=4$ branches; checkpoints in the transition region around the recovery criterion are then expanded to $K=16$.
The estimated repertoire-level recoverability is
\vspace{-0.2cm}
\begin{equation}
\widehat{V}_{\mathcal{A},\xi,B}(s_t)
=
\max_{a\in\mathcal{A}}
\widehat{Q}^{\mathrm{succ}}_{\xi,B}(s_t,a),
\end{equation}
and the empirical horizon is obtained by substituting
$\widehat{V}_{\mathcal{A},\xi,B}$ for
$V_{\mathcal{A},\xi,B}$ in Eq.~\eqref{eq:recoverability_horizon}.
The lower posterior quantile biases small-$K$ checkpoint decisions toward dismissal rather than declaring recoverability from a few favourable rollouts.

Counterfactual validity additionally requires a restored checkpoint to reproduce the same experiment as its trunk.
RESCUE therefore applies a replay-fidelity check before interpreting branch outcomes: with stochastic perturbations disabled, restoring a deterministic execution must reproduce the corresponding trunk continuation.
Fig.~\ref{fig:rescue_horizon_example} illustrates the complete measurement on one failed trajectory, including the seeded branches, raw success fractions, posterior estimate, and resulting horizon.

\subsection{Stop-and-Reset Decision Value}
\label{sec:intervention_value}

Recoverability determines whether the current attempt can still succeed, but it does not determine whether intervening is preferable to continuing. 
We therefore evaluate interventions in a common cost-to-completion space. 
A reset terminates the current attempt, incurs unit cost, and starts a fresh attempt drawn from the same workload. 
Let $c_t$ denote the execution cost per environment step and let $A_\pi$ denote the expected cost to eventual completion from such a fresh attempt.

For a branchable intervention $a\in\mathcal{A}$, let $L_a$ be the number of additional steps executed after checkpoint $s_t$. 
Its expected cost is
\begin{equation}
\footnotesize
Q^{\mathrm{cost}}_{\xi,B}(s_t,a)
=
c_t\!\left(t+\mathbb{E}_{\xi}[L_a\mid s_t,a]\right)
+
\left(1-Q^{\mathrm{succ}}_{\xi,B}(s_t,a)\right)(1+A_\pi).
\label{eq:branch_cost}
\end{equation}
For immediate stop-and-reset,
\begin{equation}
Q^{\mathrm{cost}}(s_t,\mathrm{stop})
=
c_t t + 1 + A_\pi.
\label{eq:stop_cost}
\end{equation}
The already-incurred term $c_t t$ is common to all actions at the same checkpoint and therefore cancels when comparing them. 
Thus, while $Q^{\mathrm{succ}}$ ranks branchable actions by their prospects within the current attempt, $Q^{\mathrm{cost}}$ also allows stop-and-reset to be compared on the same scale. 
All reported decision costs are interpreted as expected cost per completed task under this renewal model.

For stop-and-reset, the comparison with never intervening can be written directly from ordinary execution logs. 
Let $p$ be the failure prevalence, $d$ the fraction of failures on which the stopping rule alarms, $T_f$ the mean terminal length of those failures, and $t_d$ their mean alarm time. 
Let $f$ be the false-alarm rate on otherwise successful episodes, $t_{\mathrm{fa}}$ the mean false-alarm time, and $\bar{T}_{s,\mathrm{FA}}$ the mean natural completion length of the successful episodes that are false-alarmed. 
If $C_{\mathrm{never}}$ is the expected cost to completion when never intervening, equating the stopping rule with this baseline gives the break-even false-alarm rate
\begin{equation}
f^{*}
=
\frac{p}{1-p}
\,
\frac{
d\,c_t\left(T_f-t_d\right)
}{
\Delta+C_{\mathrm{never}}
},
\qquad
\Delta
=
1+c_t
\left(
t_{\mathrm{fa}}-\bar{T}_{s,\mathrm{FA}}
\right).
\label{eq:break_even_fa}
\end{equation}

The numerator is the expected execution cost saved by stopping detected failures early; the denominator is the excess cost of a false alarm, including the reset and replacement attempt. 
A realised false-alarm rate below $f^{*}$ makes the stopping rule preferable to never intervening under the declared cost model, whereas a larger rate makes it decision-negative. 
Setting $d=1$ and $t_d=0$ yields the maximum false-alarm rate affordable by an idealised detector that identifies every failure immediately.

\section{EXPERIMENTS}

\subsection{Experimental Setup}
\label{sec:experimental_setup}

\noindent\textit{Policies, tasks, and corpus.}
We evaluate RESCUE with OpenVLA-OFT~\cite{kim2025openvlaoft} and $\pi_{0.5}$~\cite{physicalintelligence2025pi05} on LIBERO-Spatial~\cite{liu2023libero} and its LIBERO-Plus perturbations~\cite{fei2025liberoplus}.
The complete corpus contains 1,562 trunk episodes, including 262 failures, with 23,430 saved snapshots, 10,083 branched checkpoints, and 128,340 counterfactual branches.
Our primary analysis uses the two strong-policy LIBERO-Plus cells, each containing 481 task variants.
Two weakened OpenVLA-OFT cells on LIBERO-Spatial ($150$ episodes each) and two vanilla control cells ($200$ OpenVLA-OFT and $100$ $\pi_{0.5}$ episodes) provide controlled contrasts rather than cross-policy comparisons.
All failures in the corpus terminate at the 220-step episode timeout.

\noindent\textit{Counterfactual interventions.}
Snapshots are recorded at action-chunk boundaries---every 8 environment steps for OpenVLA-OFT and every 10 for $\pi_{0.5}$---and restored in a fresh process before branching.
We instantiate the branchable repertoire as
$\mathcal{A}=\{\textsc{Continue},\textsc{Recover}\}$.
\textsc{Continue} resumes the nominal policy directly.
\textsc{Recover} executes a fixed open-loop manoeuvre before resuming the policy: freeze for 8 steps; if the trunk’s recorded gripper command at $t-1$ is positive, lower for 6 steps and release for 4; then retract for 8 steps and settle for 4.
The script therefore lasts 30 steps when this condition holds and
20 otherwise, for both policies.
The policy server is idle during the script, and success is credited only after the scripted manoeuvre has finished and nominal policy execution has resumed.
Stop-and-reset is not branched and is evaluated analytically using Sec.~\ref{sec:intervention_value}.

\noindent\textit{Execution model and recoverability criterion.}
Unless otherwise stated, counterfactual branches use an action perturbation with $\sigma=0.05$ on the six continuous arm dimensions; the gripper command and scripted recovery are not perturbed.
$\pi_{0.5}$ additionally retains its native policy sampling.
For LIBERO-Plus sensor-noise variants, branches remain conditioned on the corruption realisation observed by the corresponding trunk.
The remaining budget is inherited from the original episode,
$B(t)=220-t$, rather than restarted after intervention.
We use $\tau=0.5$ as the primary recoverability criterion and report sensitivity to $\tau\in\{0.3,0.7\}$.

\noindent\textit{Runtime monitors and decision costs.}
We evaluate four monitor families where applicable: steps-since-progress, action-jerk, sampling disagreement for $\pi_{0.5}$, and a SAFE-style implementation~\cite{gu2025safe} operating on OpenVLA-OFT hidden representations with MLP and LSTM heads.
Steps-since-progress counts action-chunk boundaries since the end-effector position last moved by more than $0.01$\,m between successive boundaries.
Action-jerk, a heuristic rather than physical jerk, is the mean absolute first difference between consecutive commanded actions over the trailing 16 steps, averaged over the six continuous arm dimensions.
For $\pi_{0.5}$, sampling disagreement is the mean standard deviation of the six arm dimensions over eight independently sampled actions at each boundary.
The SAFE-style MLP and LSTM heads use OpenVLA-OFT hidden representations as their input features.
For the primary operating point, monitor thresholds are selected on validation data to target a 5\% false-alarm rate on successful episodes and are then evaluated on held-out data using their realised test false-alarm rate.
For decision-value evaluation, one reset has unit cost and the per-step cost $c_t$ is chosen so that the median successful episode costs $0.1$ resets.
Confidence intervals for the reported decision costs use 1,000 stratified episode-level bootstrap resamples, with threshold selection repeated within each resample.

\subsection{Recoverability Is Intervention-Relative}
\label{sec:recoverability_results}

\begin{figure}[!t]
    \centering
    \includegraphics[width=0.9\columnwidth]{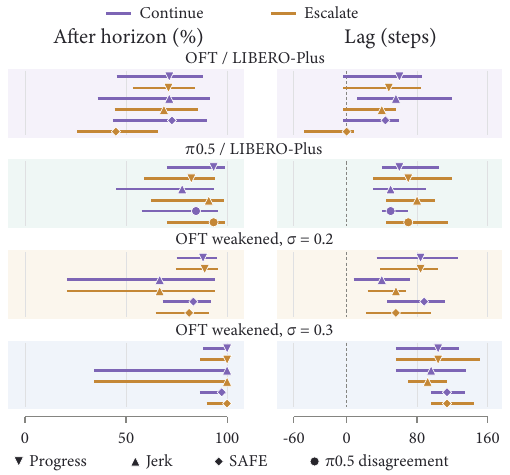}
   \vspace{-0.2cm}
   \caption{\textbf{Monitor alarms often arrive after the recoverability horizon.} 
    Across four policy settings, \textbf{left:} fraction of alarmed failures whose alarm occurs after the horizon, with 95\% Wilson confidence intervals; \textbf{right:} alarm-minus-horizon lag, with medians and interquartile ranges. Positive lag indicates late detection. Results include failures for which both an alarm and a recoverability horizon are defined under the corresponding intervention. Purple denotes \textsc{Continue}; amber denotes scripted \textsc{Recover}, labelled ``escalate'' in the figure.
   }
   \vspace{-0.4cm}
    \label{fig:recoverability_summary}
\end{figure}

We first evaluate whether recoverability is preserved under nominal continuation.
As summarised in Fig.~\ref{fig:recoverability_summary}, where the plotted last recoverable state (LRS) is the empirical recoverability horizon of Eq.~\eqref{eq:recoverability_horizon}, at $\tau=0.5$ only 14 of 66 OpenVLA-OFT/LIBERO-Plus failures contain any checkpoint satisfying the recovery criterion under
$\mathcal{A}=\{\textsc{Continue}\}$; the remaining 52 contain no qualifying checkpoint despite evaluating every saved action-chunk boundary.
Among the 14 failures with a defined horizon, the median recoverability horizon is step 44 (IQR 28--56) of the 220-step episode budget.
Relative to task-level events, this horizon occurs a median of 7 steps after the first grasp attempt and 34 steps before a subsequent object drop where those events are defined, illustrating that the horizon need not coincide with an obvious visual failure event.


Expanding the available intervention changes the result substantially.
Under scripted \textsc{Recover}, 33 of the same 66 failures contain a qualifying checkpoint, with a median horizon of step 64 (IQR 32--80).
Among the 13 failures with a defined horizon under both interventions, \textsc{Recover} shifts the horizon by a median of $+40$ steps; 11 shifts are positive and two negative (two-sided sign test, $p=0.022$).
With the combined repertoire
$\mathcal{A}=\{\textsc{Continue},\textsc{Recover}\}$,
34 of 66 failures contain at least one qualifying intervention, with a median horizon of step 64 (IQR 36--80), while 32 remain unrecoverable under either intervention.
These results directly demonstrate that recoverability depends on the intervention repertoire rather than on the robot-policy state alone.

The effect is not uniformly beneficial across conditions.
On $\pi_{0.5}$/LIBERO-Plus, 25 of 55 failures clear the criterion under each intervention individually, with median horizons of 90 steps for \textsc{Continue} and 70 for \textsc{Recover}.
On the weakened OpenVLA-OFT cells, scripted \textsc{Recover} instead reduces the clearing set from 50/51 to 43/51 at execution noise $\sigma_{\mathrm{exec}}=0.2$, and from 51/87 to 41/87 at $\sigma_{\mathrm{exec}}=0.3$.
These are within-cell intervention contrasts; we do not interpret the raw differences between OpenVLA-OFT and $\pi_{0.5}$ as a policy comparison.

Finally, the gain from scripted recovery on the primary OpenVLA-OFT cell is not solely an artefact of the injected branch perturbation.
For the 61 failures with valid deterministic zero-noise replay, \textsc{Continue} at $\sigma=0$ produces no successful continuation, whereas scripted \textsc{Recover} retains qualifying checkpoints on 31 of those failures.
Thus, changing the intervention alters the counterfactual outcome even when the injected branch perturbation is removed.

\subsection{Decision Value of Stop-and-Reset}
\label{sec:stop_reset_results}

Recoverability alone does not establish that an alarm should trigger an intervention.
We next instantiate the stop-and-reset model of Sec.~\ref{sec:intervention_value} using the observed failure prevalence, episode durations, alarm times, and realised false-alarm rates.
Table~\ref{tab:break_even} reports the complete break-even worksheet for the steps-since-progress monitor at its validation-selected nominal 5\% operating point on the two primary LIBERO-Plus cells.

The available margin for false alarms is small even before accounting for detector imperfections.
For an idealised detector with $d=1$ and $t_d=0$, Eq.~\eqref{eq:break_even_fa} yields an affordable false-alarm rate of only 2.78--3.32\% on OpenVLA-OFT/LIBERO-Plus and 2.02--2.39\% on $\pi_{0.5}$/LIBERO-Plus.
The range reflects the timing of false alarms and the natural duration of the successful episodes on which they occur.
Thus, even a detector that identifies every failure immediately can become decision-negative if it unnecessarily resets only a small fraction of otherwise successful executions.

The idealised ceiling is therefore necessary rather than sufficient.
On the weakened $\sigma_{\mathrm{exec}}=0.3$ cell, for example,
steps-since-progress realises a 9.68\% false-alarm rate within the
9.59--10.30\% perfect-detector ceiling, yet still incurs $+0.2613$
excess cost per completed task because it neither detects every
failure nor alarms at step zero.

\begin{table}[t]
\centering
\caption{
\textbf{Break-even stop-and-reset worksheet.}
Inputs to Eq.~\eqref{eq:break_even_fa} for steps-since-progress at its validation-selected nominal 5\% false-alarm operating point.
$\bar{T}_{s,\mathrm{FA}}$ is the mean natural duration of the successful episodes that are actually false-alarmed.
The realised false-alarm rate $f$ substantially exceeds the break-even rate $f^{*}$ on both primary cells.
}
\label{tab:break_even}
\resizebox{0.9\linewidth}{!}{%
\begin{tabular}{@{}llrr@{}}
\toprule
 & & OFT-LP & $\pi_{0.5}$-LP \\
\midrule
$p$ & failure prevalence & 0.1372 & 0.1143 \\
$T_f$ & failure length, steps & 220 & 220 \\
$\bar{T}_{s,\mathrm{FA}}$ & success length, false-alarmed only & 116.00 & 128.87 \\
$T_s$ & median success length, steps & 97 & 112 \\
$\kappa$ & episode time / reset & 0.1 & 0.1 \\
$C_{\mathrm{never}}$ & cost to completion, never intervene & 0.2972 & 0.2551 \\
$d$ & alarm rate on failures & 0.879 & 0.519 \\
$t_d$ & alarm step, failures & 133.0 & 129.3 \\
$t_{fa}$ & alarm step, false alarms & 61.9 & 62.7 \\
\midrule
$f^{*}$ & break-even FA rate & 1.01\% & 0.45\% \\
$f$ & \emph{realised} FA rate & 7.25\% & 7.04\% \\
\bottomrule
\end{tabular}}
\vspace{3pt}
\begin{minipage}{\columnwidth}\footnotesize $\kappa = c_t T_s$ with $T_s$ the median success length. $\bar{T}_{s,\mathrm{FA}}$ is the mean length of the successes this rule alarms on, not of all successes; with it the closed form reproduces direct episode accounting exactly.
\end{minipage}
\vspace{-0.4cm}
\end{table}

For the real steps-since-progress monitor, Table~\ref{tab:break_even} shows that the available margin is substantially smaller.
On OpenVLA-OFT/LIBERO-Plus, the monitor detects a fraction $d=0.879$ of failures at mean alarm step $t_d=133.0$, yielding a break-even false-alarm rate of only $f^{*}=1.01\%$.
Its held-out realised false-alarm rate is instead $7.25\%$.
Direct episode accounting therefore gives an excess cost of $+0.0835$ reset-equivalents per completed task relative to never intervening (95\% CI $[0.0294,\,0.1392]$).

The same pattern holds for $\pi_{0.5}$/LIBERO-Plus: Table~\ref{tab:break_even} gives $f^{*}=0.45\%$ against a realised false-alarm rate of $7.04\%$.
These results show that a nominal false-alarm target used to select a failure detector need not be economically compatible with the intervention triggered by its alarms.

\subsection{Runtime Monitors as Intervention Rules}
\label{sec:monitor_rules}

We next evaluate the runtime monitors as intervention rules rather than only as failure predictors.
The result in Sec.~\ref{sec:stop_reset_results} is not specific to steps-since-progress.
At the validation-selected nominal 5\% operating point, action-jerk has stop-and-reset excess costs of $+0.0563$ and $+0.0698$ reset-equivalents per completed task on OpenVLA-OFT/LIBERO-Plus and $\pi_{0.5}$/LIBERO-Plus, respectively.
On OpenVLA-OFT/LIBERO-Plus, the SAFE-style independent head has a three-seed mean excess cost of $+0.4988$, while the $\pi_{0.5}$ sampling-disagreement monitor is closest to the baseline at $+0.0235$.
None of the evaluated monitor families therefore provides evidence of lower cost than never intervening on the primary strong-policy conditions.
This is not a universal no-intervention result.
On the higher-prevalence weakened conditions, the SAFE-style monitor has lower cost than never intervening on three of six held-out evaluations; the largest gain is $\Delta C=-0.0431$ (95\% CI $[-0.0432,\,-0.0426]$) on a cell with 34\% failure prevalence.

Changing the intervention triggered by the same alarm can improve the outcome without making the monitor beneficial overall.
Routing alarms to scripted \textsc{Recover} instead of immediate stop-and-reset reduces excess cost by factors of 3.1--13.4 across the evaluated monitor configurations.
Nevertheless, on the two primary strong-policy conditions, no real monitor has a confidence interval establishing lower cost than never intervening under either intervention rule.
Thus, the value of an alarm depends on the intervention it triggers, even when the monitor itself is unchanged.

To test whether this result is caused by the nominal 5\% operating point, we also select each threshold on validation data by minimising intervention cost.
Of the eight monitor--intervention configurations, five select thresholds that produce no alarms on the held-out set.
The only negative point estimate, steps-since-progress routed to \textsc{Recover}, gives $\Delta C=-0.0058$, but its 95\% interval $[-0.0204,\,0.0143]$ spans zero.
Selecting the operating point directly for intervention cost therefore does not establish an improvement over never intervening.

\begin{table}[!t]
\centering
\caption{
\textbf{Detection/timing metrics and intervention value can disagree.}
Entries report the number of evaluation universes admitting each monitor.
Across the real monitors, detection/timing criteria admit 61/75 verdicts, while decision value admits 0/15 against never intervening and 5/15 against rate-matched random alarming.
$\dagger$ denotes decoys.
}
\label{tab:metric_verdicts}
\resizebox{0.99\linewidth}{!}{%
\begin{tabular}{l c c c c c c c c}
\toprule
& &
\multicolumn{5}{c}{\textbf{Detection / timing metrics}} &
\multicolumn{2}{c}{\textbf{Decision value}} \\
\cmidrule(lr){3-7}
\cmidrule(lr){8-9}
\textbf{Monitor} &
$\mathbf{u}$ &
\textbf{End} &
\textbf{Early} &
\textbf{$<100$} &
\textbf{AUCPDT} &
\textbf{Timely} &
\textbf{vs.\ Never} &
\textbf{vs.\ Random} \\
\midrule
Steps-since-progress        & 4 & 4/4 & 1/4 & 2/4 & 4/4 & 2/4 & 0/4 & 3/4 \\
Action-jerk                 & 4 & 4/4 & 1/4 & 2/4 & 4/4 & 2/4 & 0/4 & 1/4 \\
SAFE-style MLP              & 3 & 3/3 & 3/3 & 3/3 & 3/3 & 3/3 & 0/3 & 0/3 \\
SAFE-style LSTM             & 3 & 3/3 & 3/3 & 3/3 & 3/3 & 3/3 & 0/3 & 0/3 \\
Sampling disagreement       & 1 & 1/1 & 1/1 & 1/1 & 1/1 & 1/1 & 0/1 & 1/1 \\
\midrule
Fixed-step clock$^\dagger$  & 4 & 4/4 & 1/4 & 3/4 & 3/4 & 3/4 & 1/4 & 3/4 \\
Label leak$^\dagger$        & 4 & 4/4 & 4/4 & 4/4 & 4/4 & 4/4 & 4/4 & 4/4 \\
$-$Steps-since-progress$^\dagger$
                            & 4 & 0/4 & 0/4 & 0/4 & 0/4 & 0/4 & 0/4 & 4/4 \\
Random alarm$^\dagger$      & 4 & --  & --  & --  & --  & 0/4 & 0/4 & 0/4 \\
\bottomrule
\end{tabular}}
\vspace{-0.4cm}
\end{table}

We next test whether conventional monitor-selection metrics lead to the same verdict.
Table~\ref{tab:metric_verdicts} compares the real monitors on matched evaluation universes using three AUROC variants, VLA-FAIL's AUCPDT~\cite{seligmann2026vlafail}, a timing criterion, and decision value.
\emph{End} is episode-level AUROC using the maximum monitor score over the full trajectory; \emph{Early} restricts that maximum to the task-specific early-exposure window used by SAFE; and \emph{$<100$} restricts it to checkpoints before environment step 100.
\emph{Timely} is the horizon-relative lead rate at the 5\% operating point: the fraction of alarmed failures with a defined recoverability horizon whose alarm occurs at or before that horizon.
For this comparison, confidence intervals are computed by resampling the underlying base tasks to account for dependence among LIBERO-Plus variants.
A metric ``admits'' a monitor when its 95\% base-task interval excludes the corresponding reference in the favourable direction; for decision value, this requires lower cost than the reference.
Across the 15 real-monitor--universe combinations, the five detection and timing criteria admit the monitors in 61 of 75 possible verdicts.
Decision value, in contrast, admits them in 0 of 15 verdicts against never intervening.

The reference used for decision value also affects the verdict.
Against a rate-matched random alarm, decision value admits 5 of 15 real-monitor verdicts; however, the deliberately inverted steps-since-progress decoy also clears this reference on all four universes.
Outperforming a random alarm is therefore not sufficient to establish useful intervention value.
The 61/75 and 0/15 counts are descriptive tallies over related metrics and overlapping evaluation sets, not independent statistical tests.

A sensor-free fixed-step baseline provides another reference.
On OpenVLA-OFT/LIBERO-Plus, a validation-selected fixed-step rule stops at step 168 and costs $+0.0038$ reset-equivalents per completed task (95\% CI $[-0.0085,\,0.0348]$), compared with $+0.0835$ for steps-since-progress.
Failure discrimination alone therefore does not determine intervention value.

The monitor families fail for different reasons.
Steps-since-progress alarms late relative to the recoverability horizon, limiting the execution time that stopping can save.
The SAFE-style monitor ranks failures effectively but transfers its nominal operating point poorly: the independent head reaches a three-seed mean realised false-alarm rate of 26.5\% at a 5\% validation target.
Conversely, $\pi_{0.5}$ sampling disagreement realises a 2.3\% false-alarm rate---within the idealised ceiling of Sec.~\ref{sec:stop_reset_results}---yet still has positive estimated intervention cost.
Discrimination, false-alarm rate, and intervention timing therefore capture different aspects of monitor usefulness.

\subsection{Robustness and Evaluation Controls}
\label{sec:robustness_controls}

\noindent\textit{Counterfactual fidelity and estimator stability.}
The counterfactual analysis is meaningful only if restoring a checkpoint reproduces the experiment from which it was saved.
Across all six cells, 3,000/3,000 seeded checkpoint replays reproduced the recorded joint trajectory with maximum joint-position deviation $0.0$.
A stronger zero-noise known-answer test also exposed an observation-channel mismatch on the LIBERO-Plus sensor-noise track; after correcting the branch runner, the affected episodes were re-branched and all results reported here use the corrected labels.
We separately re-evaluated 20 OpenVLA-OFT/LIBERO-Plus failures with a fixed $K=32$ at every checkpoint of both interventions (35,840 branches).
At $\tau=0.5$, the published estimator disagrees with this higher-sampling reference on only 9/560 \textsc{Continue} checkpoints (1.61\%) and 6/560 \textsc{Recover} checkpoints (1.07\%).
Whether an episode contains any qualifying checkpoint is unchanged for all 20 failures under both interventions; the estimated horizon moves on 3/20 episodes for \textsc{Continue} and 1/20 for \textsc{Recover}.
We therefore interpret the fixed-$K$ experiment as agreement under increased sampling rather than as ground truth.
The same fixed-$K$ subset also shows that recoverability need not evolve monotonically with execution time.
Against a monotone null, \textsc{Recover} is significantly non-monotone on 8/20 episodes, whereas \textsc{Continue} is non-monotone on 0/20 at this sampling level.
This supports defining the horizon as the latest qualifying checkpoint without assuming a monotone recoverability boundary.

\noindent\textit{Decision-cost robustness.}
Because LIBERO-Plus variants are grouped around ten underlying base tasks, episode-level resampling may understate uncertainty when variants from the same task are correlated.
We therefore repeated the decision-cost analysis using base-task rather than episode resampling.
For steps-since-progress on OpenVLA-OFT/LIBERO-Plus, the 95\% interval widens from $[0.0294,\,0.1392]$ to $[0.0234,\,0.2793]$ but remains above zero.
The zero-exclusion verdict is unchanged across the scored monitor configurations.
The stop-and-reset conclusion is also not specific to the chosen time price.
When a median successful episode is valued at $0.05$, $0.1$, $0.2$, or $0.5$ resets, steps-since-progress on OpenVLA-OFT/LIBERO-Plus remains decision-negative, with excess costs of $+0.0870$, $+0.0835$, $+0.0764$, and $+0.0551$, respectively.

\noindent\textit{Sensitivity to the recoverability definition.}
The main conclusions are not tied to a single recovery threshold.
For OpenVLA-OFT/LIBERO-Plus under \textsc{Continue}, 18, 14, and 13 of 66 failures contain a qualifying checkpoint at $\tau=0.3$, $0.5$, and $0.7$, respectively, with median horizons at steps 40, 44, and 32.
The clock convention can also change the answer: replacing the inherited deadline with a fresh budget equal to the median successful-episode length shifts the paired $\pi_{0.5}$/LIBERO-Plus horizon by $+30$ steps (IQR $0$--$55$) on the 27 failures labelled under both conventions and changes 29/55 episode labels.
Because this auxiliary clock ablation was run before the LIBERO-Plus sensor-noise branch correction, we use it only to assess sensitivity to $B$; all primary horizon results use the corrected labels.

\noindent\textit{Timing requires a matched null.}
A monitor alarming after the recoverability horizon does not by itself establish poor timing because the horizon often closes early in these trajectories.
A uniformly random alarm also occurs after the horizon on approximately 73\% of recoverable failures in each OpenVLA-OFT cell.
Indeed, the OpenVLA-OFT/LIBERO-Plus SAFE-style timing rates fall within the corresponding random-alarm interval.
We therefore compare timing metrics with alarms generated on the same episode population rather than treating zero lag as an implicit null.
Nevertheless, under \textsc{Continue}, every real monitor alarms after the horizon in more than half of the episodes for which both quantities are defined.

\noindent\textit{Timeout-length confounding.}
All 262 failures in the corpus are 220-step timeouts, creating a strong length cue for episode-level detection metrics.
A score using only the number of observed action-chunk boundaries attains AUROC $1.000$ on all 18 validation splits and again on all three task-disjoint holdouts.
Using a per-task early-exposure control reduces the steps-since-progress AUROC from $0.979$ to $0.637$, demonstrating that uncorrected episode-level discrimination can substantially reflect trajectory length rather than failure information.
We therefore treat the conventional detection metrics in Table~\ref{tab:metric_verdicts} as monitor-selection diagnostics, not as substitutes for intervention-value evaluation.

\section{DISCUSSION AND LIMITATIONS}
\label{sec:discussion}

RESCUE shifts runtime-monitor evaluation from the monitor alone to the monitor--intervention pair.
A useful alarm must arrive while an effective intervention remains available and while acting is preferable under the prevalence and costs of the intended workload.
For stop-and-reset, Eq.~\eqref{eq:break_even_fa} makes this comparison directly from ordinary execution logs, using the monitor's realised held-out false-alarm rate rather than its nominal target.
Recoverability is likewise conditional on the declared repertoire $\mathcal{A}$, execution model $\xi$, and remaining budget $B$; it should therefore not be interpreted as an intrinsic property of a robot state or policy.
A different recovery controller, deadline, execution model, or cost structure can change both the recoverable set and the preferred intervention.
Here we evaluate only \textsc{Continue} and one fixed scripted \textsc{Recover}, and make no claim about richer or learned recovery repertoires.

Several limitations bound the empirical conclusions.
All experiments use MuJoCo and LIBERO/LIBERO-Plus, and all 262 observed failures are timeouts; hardware may introduce contact dynamics, damage, latency, and recovery costs absent from this benchmark.
The branch perturbation and decision costs are declared rather than calibrated from deployment.
We evaluate two VLA policies and four monitor families, without training a new detector.
The lightweight-monitor splits are episode-disjoint but not base-task-disjoint, so their false-alarm rates do not establish generalisation to unseen tasks.
Base-task uncertainty estimates also use few clusters---ten overall and three held-out base tasks per SAFE-style seed---and may therefore be optimistic.
The two LIBERO-Plus policy cells fail on different perturbation tracks, precluding direct policy comparison.
Finally, the fixed-$K$ analysis in Sec.~\ref{sec:robustness_controls} supports episode-level stability but does not make the estimated horizon noise-free.

\section{CONCLUSION}
\label{sec:conclusion}

We introduced RESCUE, a counterfactual framework for evaluating runtime monitors together with the interventions their alarms trigger.
Rather than treating failure prediction as sufficient, RESCUE measures action-conditioned recoverability from restored robot-policy states and evaluates interventions in a common cost-to-completion space.
Across 1,562 VLA episodes and 128,340 counterfactual branches, we find that recoverability changes with the available intervention and that monitors favoured by conventional detection and timing metrics can remain decision-negative once false alarms, timing, prevalence, and intervention cost are included.
These results support evaluating runtime monitors together with the interventions they trigger, rather than relying on failure-prediction quality alone.



\bibliographystyle{IEEEtran}
\bibliography{refs}

\end{document}